\documentclass[11pt,a4paper]{article}

\usepackage{expex}
\usepackage{mathtools}
\usepackage{booktabs}
\usepackage{tabularx}
\usepackage{hyperref}
\usepackage{xcolor}		
  \definecolor{darkblue}{rgb}{0, 0, 0.5}
  \hypersetup{colorlinks=true,citecolor=darkblue, linkcolor=darkblue, urlcolor=darkblue}
  
\lingset{
    textoffset=.5em,
    labeloffset=.5em
}

\let\tok\textsc
\def\lex#1{\emph{#1}}
\def\begnobreak{\par\nobreak\vfil\penalty0\vfilneg\vtop\bgroup}
\def\endnobreak{\par\xdef\tpd{\the\prevdepth}\egroup\prevdepth=\tpd}
\let\P\relax

\DeclareMathOperator{\alice}{\tok{alice}}
\DeclareMathOperator{\bob}{\tok{bob}}
\DeclareMathOperator{\john}{\tok{john}}
\DeclareMathOperator{\mary}{\tok{mary}}
\DeclareMathOperator{\student}{\tok{student}}
\DeclareMathOperator{\teacher}{\tok{teacher}}
\DeclareMathOperator{\see}{\tok{see}}
\DeclareMathOperator{\run}{\tok{run}}
\DeclareMathOperator{\Verb}{\emph{verb}}
\DeclareMathOperator{\P}{\emph{P}}

\usepackage{coling2020}
\usepackage{mathptmx}
\usepackage{latexsym}

\usepackage{natbib}
\renewcommand\cite{\citep}	
\usepackage{microtype}

\colingfinalcopy 

\makeatletter
\newcommand{\printfnsymbol}[1]{%
  \textsuperscript{\@fnsymbol{#1}}%
}
\makeatother

\title{Sequence-to-Sequence Networks \\ Learn the Meaning of Reflexive Anaphora}

\author{Robert Frank\thanks{\hspace{0.2cm}Equal contribution} \\
  Yale University \\
  \texttt{bob.frank@yale.edu} \\\And
  Jackson Petty\printfnsymbol{1} \\
  Yale University \\ 
  \texttt{jackson.petty@yale.edu} \\}
\date{\today}

\begin{document}
\maketitle
\begin{abstract}
Reflexive anaphora present a challenge for semantic interpretation: their meaning varies depending on context in a way that appears to require abstract variables. Past work has raised doubts about the ability of recurrent networks to meet this challenge. In this paper, we explore this question in the context of a fragment of English that incorporates the relevant sort of contextual variability. We consider sequence-to-sequence architectures with recurrent units and show that such networks are capable of learning semantic interpretations for reflexive anaphora which generalize to novel antecedents. We explore the effect of attention mechanisms and different recurrent unit types on the type of training data that is needed for success as measured in two ways: how much lexical support is needed  to induce an abstract reflexive meaning (i.e., how many distinct reflexive antecedents must occur during training) and what contexts must a noun phrase occur in to support generalization of reflexive interpretation to this noun phrase?
\end{abstract}

\section{Introduction}

Recurrent neural network architectures have demonstrated remarkable success in natural language processing, achieving state of the art performance across an impressive range of tasks ranging from machine translation to semantic parsing to question answering \citep{sutskever2014sequence, gru, bahdanau2016neural}. These tasks demand the use of a wide variety of computational processes and information sources (from grammatical to lexical to world knowledge), and are evaluated in coarse-grained quantitative ways. As a result, it is not an easy matter to  identify the specific strengths and weaknesses in a network's solution of a task.  

In this paper, we take a different tack, exploring the degree to which neural networks successfully master one very specific aspect of linguistic knowledge: the interpretation of sentences containing reflexive anaphora.  We address this problem in the context of the task of semantic parsing, which we instantiate as mapping a sequence of words into a predicate calculus logical form representation of the sentence's meaning.
\pex<ex:transform>
    \a Mary runs $\to$ $\run(\mary)$
    \a John sees Bob $\to$ $\see(\john, \bob)$
\xe
Even for simple sentences like those in~(\getref{ex:transform}), which represent the smallest 
representations of object reflexives in English, the network must learn lexical semantic 
correspondences (e.g., the input symbol \lex{Mary} is mapped to the output $\mary$ and  \lex{runs}
is mapped to $\run$) and a mode of composition (e.g., for an intransitive sentence, the meaning of the subject is surrounded by parentheses and appended to the meaning of the verb).  
Of course, not all of natural language adheres to such simple formulas. Reflexives, words like {\em herself} and {\em himself}, do not have an interpretation that can be assigned independently of the meaning of the surrounding context.
\pex<ex:transform-refl>
    \a Mary sees herself $\to \see(\mary, \mary)$
    \a Alice sees herself $\to \see(\alice, \alice)$
\xe
In these sentences, the interpretation of the reflexive is not a constant that can be combined with the meaning of the surrounding elements. Rather, a reflexive object must be interpreted as identical to the meaning of verb's subject.  Of course, a network could learn a context-sensitive interpretation of a reflexive, so that for any sentence with \lex{Mary} as its subject, the reflexive is interpreted as $\mary$, and with \lex{Alice} as its subject it is interpreted as $\alice$.  However, such piecemeal learning of reflexive meaning will not support generalization to sentences involving a subject that has not been encountered as the antecedent of a reflexive during training, even if the interpretation of the  subject has occurred elsewhere. What is needed instead is an interpretation of the reflexive that is characterized not as a specific (sequence of) output token(s), but rather as an abstract instruction to duplicate the interpretation of the subject. Such an abstraction requires more than the ``jigsaw puzzle" approach to meaning that simpler sentences afford. 

\citet{Marcus98} argues that this kind of abstraction, which he takes to require the use of  algebraic variables to assert identity, is beyond the capacity of recurrent neural networks. 
\citeauthor{Marcus98}'s demonstration involves a simple recurrent network (SRN, \citealt{elman90}) language model that is trained to predict the next word over a corpus of sentences of the following form:
\pex
    \a A rose is a rose.
    \a A mountain is a mountain.
\xe
All sentences in this training set have identical subject and object nouns. 
\citeauthor{Marcus98} shows, however, that the resulting trained network does not correctly predict the subject noun when tested with a novel preamble `\lex{A book is a $\ldots$}'. Though intriguing, this demonstration is not entirely convincing: since the noun occurring in the novel preamble, \lex{book} in our example, did not occur in the training data, there is no way that the network could possibly have known which (one-hot represented) output should correspond to the reflexive for a sentence containing the novel (one-hot represented) subject noun, even if the network did successfully encode an identity relation between subject and object. 

\citet{frank2013} explore a related  task in the context of SRN interpretation of reflexives. In their experiments, SRNs were trained to map input words to corresponding semantic symbols that are output on the same time step in which a word is presented. For most words in the vocabulary, this is a simple task: the desired output is a constant function of the input (\lex{Mary} corresponds to $\mary$, \lex{sees} to \textsc{see},  etc.).  For reflexives however, the target output depends on the subject that occurs earlier in the sentence. \citeauthor{frank2013}\ tested the network's ability to interpret a reflexive in sentences containing a subject that had not occurred as a reflexive's antecedent during training. However, unlike Marcus' task, this subject and its corresponding semantic symbol did occur in other (non-reflexive) contexts in the training data, and therefore was in the realm of possible inputs and outputs for the network. Nonetheless, none of the SRNs that they trained succeeded at this task for even a single test example.

Since those experiments were conducted, substantial advances have been made on recurrent neural network architectures, some of which have been crucial in the success of practical NLP systems. 
\begin{itemize}
    \item\textbf{Recurrent units}: More sophisticated recurrent units like LSTMs \citep{lstm} and GRUs \citep{gru} have been shown to better encode preceding context than SRNs.
    \item \textbf{Sequence-to-Sequence architectures}: The performance of network models that transduce one string to another, used in machine translation and semantic parsing, has been greatly improved by the use of independent encoder and decoder networks  \cite{sutskever2014sequence}.
    \item \textbf{Attention mechanism}: The ability of a network to produce contextually appropriate outputs even in the context of novel vocabulary items has been facilitated by content-sensitive attention mechanisms \cite{bahdanau2016neural, luong-etal-2015-effective}.  
\end{itemize}
These innovations open up the possibility that modern network architectures may well be able to solve the variable identity problem necessary for mapping reflexive sentences to their logical form. In the experiments we describe below, we explore whether this is the case.

\section{Experimental Setup}

Our experiments take the form of a semantic parsing task, where sequences of words are mapped into logical form representations of meaning. Following \citet{dong-lapata-2016-language}, we do this by means of a sequence-to-sequence architecture  \cite{sutskever2014sequence} in which the input sentence is fully processed by an encoder network before it is decoded into a sequence of symbols in the target domain (cf.\ \citealt{BotvinickPlaut07}, \citealt{FrankMathis07} for antecedents). This approach removes the need to synchronize the production of output symbols with the input words, as in \citet{frank2013}, allowing greater flexibility in the nature of semantic representations.

The sequence-to-sequence architecture is agnostic as to the types of recurrent units for the encoding and decoding phases of the computation, and whether the decoder makes use of an attention mechanism. Here, we explore the effects of using different types
of recurrent units and including attention or not. Specifically, we examine the performance
and training characteristics of sequence-to-sequence models based on SRNs, GRUs, and LSTMs with and without multiplicative attention \cite{luong-etal-2015-effective}.

In all experiments, we perform 5 runs with different random seeds for each  combination of recurrent unit type (one layer of SRN, LSTM or GRU units for both the encoder and decoder) and attention (with or without multiplicative attention).  All models used hidden and embedding of size of 256.  Training was done using Stochastic Gradient Descent with learning rate of $0.01$. Models were trained for a maximum of 100 epochs with early stopping when validation loss fails to decrease by $0.005$ over three successive epochs.

We conduct all of our experiments with synthetic datasets from a small fragment of English sentences generated using a simple  context-free grammar. This fragment includes simple sentences with transitive and intransitive verbs. Subjects are always proper names and objects are either proper names or a reflexive whose gender matches that of the subject.  Our vocabulary includes 8 intransitive verbs, 7 transitive verbs, 15 female names, and 11 male names. The grammar thus generates 5,122 distinct sentences. All sentences are generated with equal probability, subject to the restrictions imposed by each experiment.  We use a unification extension to CFG to associate each sentence with a predicate calculus interpretation. The symbols corresponding to the predicates  and the entities in our logical language are identical with the verbs and names used by our grammar, yielding representations like those shown in (\getref{ex:transform}) and (\getref{ex:transform-refl}). The output sequences corresponding to the target semantic interpretations include parentheses and commas as separate symbols. Quite clearly, this dataset does not reproduce the richness of English sentence structure or the distribution of reflexive anaphora, and we leave the exploration of syntactically richer domains for future work. However, even this simple fragment instantiate the kind of contextual variable interpretation found in all cases of reflexive interpretation and therefore it allows us to probe the ability of networks to induce a representation of such meanings.

As discussed in the previous section, we are interested in whether sequence-to-sequence models  can successfully \emph{generalize} their knowledge of the interpretation of sentences containing reflexives to ones having novel
antecedents. 
To do this, we employ a \emph{poverty of the stimulus} 
paradigm that tests for systematic generalization beyond a finite (and ambiguous) set of training data \cite{rules}.  In our experiments, we remove certain classes of examples from the training data set  and test the effect on the network's success in interpreting reflexive-containing sentences. Each of our experiments thus defines a set of sentences that are withheld during training. The non-withheld sentences are randomly split $80\%$--$10\%$--$10\%$ between training, validation, and testing sets. Accuracy for each set is computed on a sentence-level basis, i.e., an accurate output requires that all symbols generated by the model be identical to the target. 
Our experiments focus on two sorts of manipulations of the training data: (1) varying the number of lexical items that do and do not occur as the antecedents of reflexives in the training set, and (2) varying the syntactic positions in which the non-antecedent names occur.  As we will see, both of these manipulations  substantially impact the success of reflexive generalization in ways that vary across network types.

\section{Experiment 1: Can Alice know herself?}
In the first experiment, we directly test whether or not networks can generalize
knowledge of how to interpret \lex{herself} to a new antecedent. We withhold all examples whose input sequence includes the reflexive \lex{herself} bound by the single antecedent \lex{Alice},  of the form shown
in~(\nextx).
\ex<ex:alice-herself>
    Alice \emph{verbs} herself $\to$ $\Verb(\alice, \alice)$
\xe
Sentences of any other form are included in the training-validation-test splits, including those where 
\lex{Alice} appears without binding a reflexive. 

\subsection{Results}
All network architectures were successful in this task, generalizing the 
interpretation of \lex{herself} to the novel antecedent \lex{Alice}. 
Even the simplest networks, namely SRN models without attention, achieve 100\% accuracy on the generalization set (sentences of the
form shown  in~(\getref{ex:alice-herself})). This is in sharp contrast
the negative results obtained by \citet{frank2013}, suggesting an advantage
for training with a language with more names as well as for instantiating
the semantic parsing task in a sequence-to-sequence architecture as opposed to a language
model.

\section{Experiment 2: Doesn't Alice know Alice?}
While the networks in Experiment 1 are not trained on sentences of the form shown in~(\getref{ex:alice-herself}), they are trained on sentences that have the same target semantic form, namely sentences in which \lex{Alice} occur as both subject and object of a transitive verb. 
\ex<ex:alice-alice>
    Alice \emph{verbs} Alice $\to$ $\Verb(\alice, \alice)$
\xe
In Experiment 2 we consider whether the presence of such semantically reflexive forms in the training data is helpful to networks in generalizing to syntactically reflexive sentences.  We do this by further excluding sentences of the form in~(\getref{ex:alice-alice}) from the training  data.

\subsection{Results}
All architectures except SRNs without attention 
generalize perfectly to the held out items. Inattentive SRNs also
generalize quite well, though only at a mean accuracy of 86\%. 
While success at Experiment 1 demonstrates the networks' abilities
to generalize to novel input contexts, success at Experiment~2
highlights how models can likewise generalize to produce 
entirely new outputs.

\section{Experiment 3: Who's Alice and who's Claire?}
So far, we have considered generalization of reflexive interpretation
to a single new name. One possible explanation of the networks' success is that they are simply 
defaulting to the (held-out) $\alice$ interpretation when confronted with a new antecedent,
as an elsewhere interpretation (but see \citealt{gandhi2019mutual} 
for reasons for skepticism).  Alternatively, even if the network has acquired
a generalized interpretation for reflexives, it may be possible that this 
happens only when the training data includes overwhelming lexical support 
(in Experiments 1 and 2, 25 out of the 26 names in our domain appeared in 
the training data as the antecedent of a reflexive).  To explore the 
contexts under which networks can truly generalize to a range of new 
antecedents, we construct training datasets in which we progressively
withhold more and more names in sentences of the forms shown in~(\nextx), i.e., those 
that were removed in Experiment 2.\footnote{Since \lex{himself} and 
\lex{herself} are different lexical items, it is unclear if the network will
learn their interpretations together, and whether sentences containing \lex{himself} 
will provide support for the interpretation of sentences containing \lex{herself}. We therefore  withhold only sentences of this form with names of a single gender. We have also experimented with witholding masculine reflexive antecedents from the training data, but the main effect remains the number of female antecedents that is withheld.}  
\pex<ex:gen-forms>
    \a $\P$ \emph{verbs} herself $\to$ $\Verb(\P,\P)$
    \a $\P$ \emph{verbs} $\P$ $\to$ $\Verb(\P,\P)$
\xe
Our domain contains 15 distinct feminine antecedents; we
perform several iterations of this experiment, withholding progressively
more feminine names from appearing in the contexts 
in~(\getref{ex:gen-forms}), until only a single feminine name is included in the training data as the antecedent of a reflexive.

\subsection{Results}

\begin{figure}[!t]
    \centering
    \includegraphics[width=0.7\columnwidth]{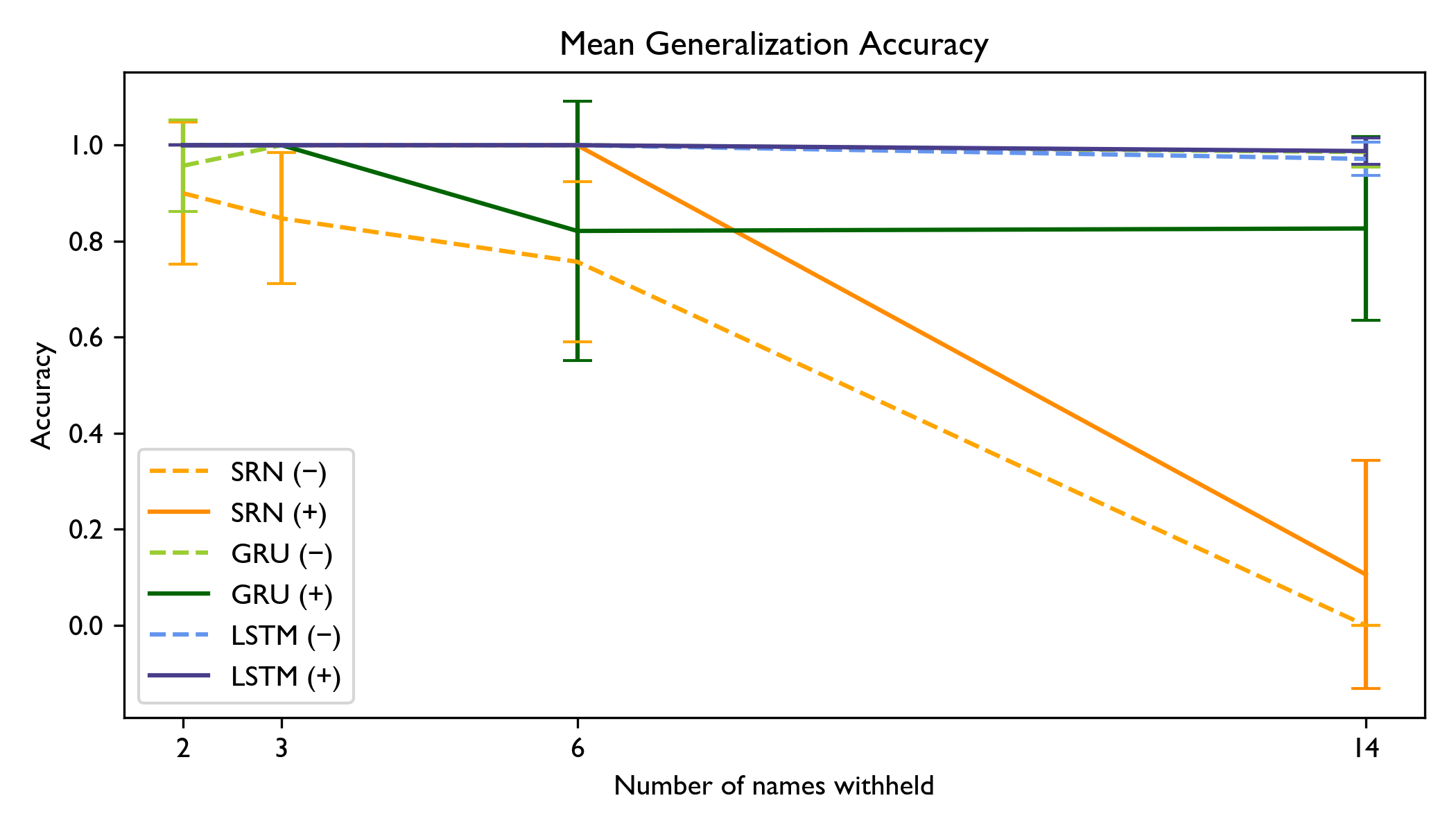}
     \vspace*{-1.5em}
    \caption{Mean generalization accuracy by number of names withheld in Experiment~3. The ($+$) or ($-$) next to the type of recurrent unit indicates the presence or absence of attention.
    Error bars display the standard deviation of accuracies. }
    \label{fig:three-names}
\end{figure}

As shown in Figure~\ref{fig:three-names}, reducing the set of names that serve as antecedents to reflexives in the training data resulted in lower accuracy on the generalization set.
SRNs, especially without attention, show significantly degraded performance when
high numbers of names are withheld from reflexive contexts during training. With attention,  SRN performance degrades only when reflexives are trained with a single feminine antecedent (i.e., 14 names are held out). In contrast, LSTMs both with and without attention maintain near-perfect accuracy on the
generalization set even when the training data allows only a single antecedent for the feminine reflexive \lex{herself}.  The performance of GRUs varies with the presence of an attention mechanism: without attention, GRUs achieve near perfect generalization accuracy even for the most demanding case (training with a single feminine antecedent), while the performance of GRUs with attention has mean accuracy near 80\%.

We also
explored how recurrent unit type and attention affect \emph{how} models
learn to generalize. One way to gauge this is by examining how quickly
networks go from learning reflexive interpretation for a single name to 
learning it for every name. 
Table~\ref{tab:learning-deltas} shows the mean number of epochs it takes
from when a network attains 95\% accuracy on a single antecedent 
contexts\footnote{An `antecedent context' is the set of all reflexive
sentences with a particular antecedent.}
to when it has attained more than 95\% accuracy on \emph{all} held out 
antecedent contexts.\footnote{Note that this doesn't mean that
models retained more than 95\% accuracy on all contexts --- some models learned
a context, only to forget it later in training; this measurement does not
reflect any such unlearning by models.}

\begin{figure}[!t]
    \centering
    \includegraphics[width=0.7\columnwidth]{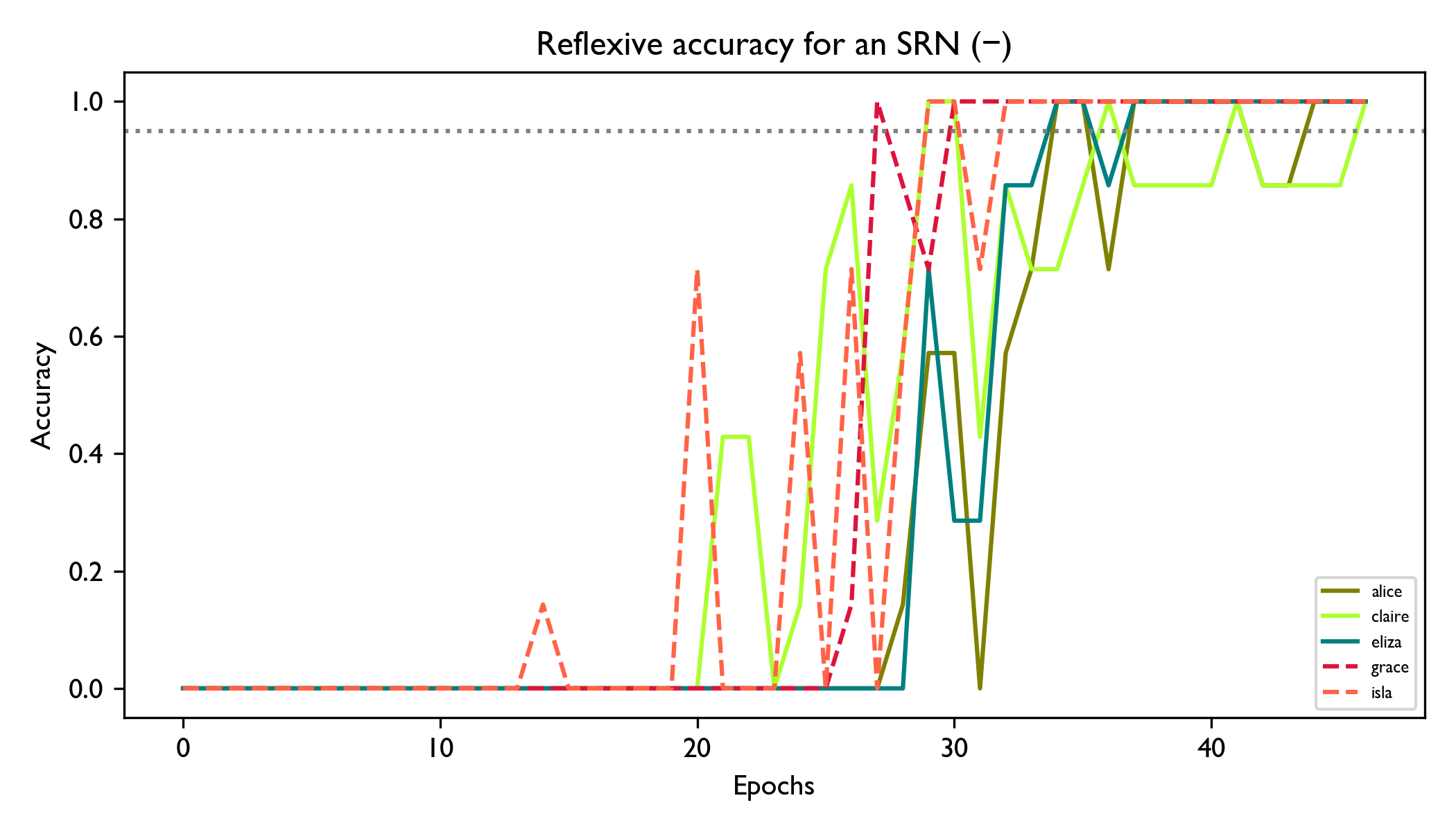}
     \vspace*{-2em}
    \caption{Reflexive accuracy with different antecedents during training of an SRN without attention. \lex{Alice}, \lex{Claire} and \lex{Eliza} were withheld during training while 
    \lex{Grace} and \lex{Isla}
    present in the training data.}
    \label{fig:all-names}
\end{figure}

\begin{table}[!t]
    \centering
    \begin{tabularx}{\columnwidth-2in}{Xrrrr} \toprule
   \textit{Architecture}  & \multicolumn{4}{c}{\emph{\# contexts withheld}}\\
         & 2 & 3 & 6 & 14 \\ \midrule
        SRN ($-$) & 7.5 & 5.0 & --- & --- \\
        SRN ($+$) & 0.6 & 0.6 & 0.6 & --- \\
        GRU ($-$) & 1.8 & 2.2 & 3.4 & 9.4 \\
        GRU ($+$) & 2.2 & 3.6 & 5.3 & 1.5 \\
        LSTM ($-$) & 1.2 & 2.2 & 4.4 & 12.2 \\
        LSTM ($+$) & 0.6 & 0.8 & 1.4 & 3.4 \\ \bottomrule
    \end{tabularx}
    \caption{Average number of epochs between having learned one context and
    having learned all contexts, calculated as the mean difference among runs
    which succeeded in eventually learning all contexts once. A `---' in a row indicates that no models  were able to achieve this degree of generalization.}
    \label{tab:learning-deltas}
\end{table}
This `time to learn' highlights the disparate impact of attention depending
on the type of recurrent unit; SRNs with attention and LSTMs with attention
acquire the generalization much faster than their attentionless counterparts,
while attention increases the length of time it takes for GRUs
to learn for all but the condition in which 14 antecedents were withheld. Figure~\ref{fig:all-names} illustrates another important aspect of reflexive generalization: it proceeds in a piecemeal fashion, where
networks first learn to interpret reflexives for the trained names and then generalize to the held out antecedents one by one. 
In Figure~\ref{fig:all-names} we show an SRN without attention, but the same pattern is representative of
the other networks tested.

\section{Experiment 4: What if Alice doesn't know anyone?}

The experiments we have described thus far removed from the training data input sentences and logical forms that were exactly identical to those associated with reflexive sentences.  The next pair of experiments increases the
difficulty of the generalization task still further,  by withholding from the Experiment 2 training data all sentences containing the withheld reflexive antecedent, \lex{Alice}, in a wider range of grammatical
contexts, and testing the effect that this has on the network's ability to interpret \lex{Alice}-reflexive sentences.

Experiment 4a starts by withholding  sentences where \lex{Alice} appears as
the subject of a transitive verb (including those with reflexive objects, which we already removed in earlier experiments). 
This manipulation tests the degree to which the presence of \lex{Alice} as a subject more generally is crucial to the network's generalization of reflexive sentences to a novel name.  We also run a variation of this experiment (Experiment 4b) in which sentences containing \lex{Alice} as the subject of intransitives are also removed, i.e., sentences of the following form:
\pex
    Alice \emph{verbs} $\to$ $\Verb(\alice)$
\xe
If subjecthood is represented in a uniform manner across transitive and intransitive sentences, the absence of such sentences from the training data might further impair the network's ability to generalize to reflexive sentences.

\pagebreak
\subsection{Results}

\paragraph{Experiment 4a}

\begin{figure*}[!th]
    \centering
    \includegraphics[width=0.45\columnwidth]{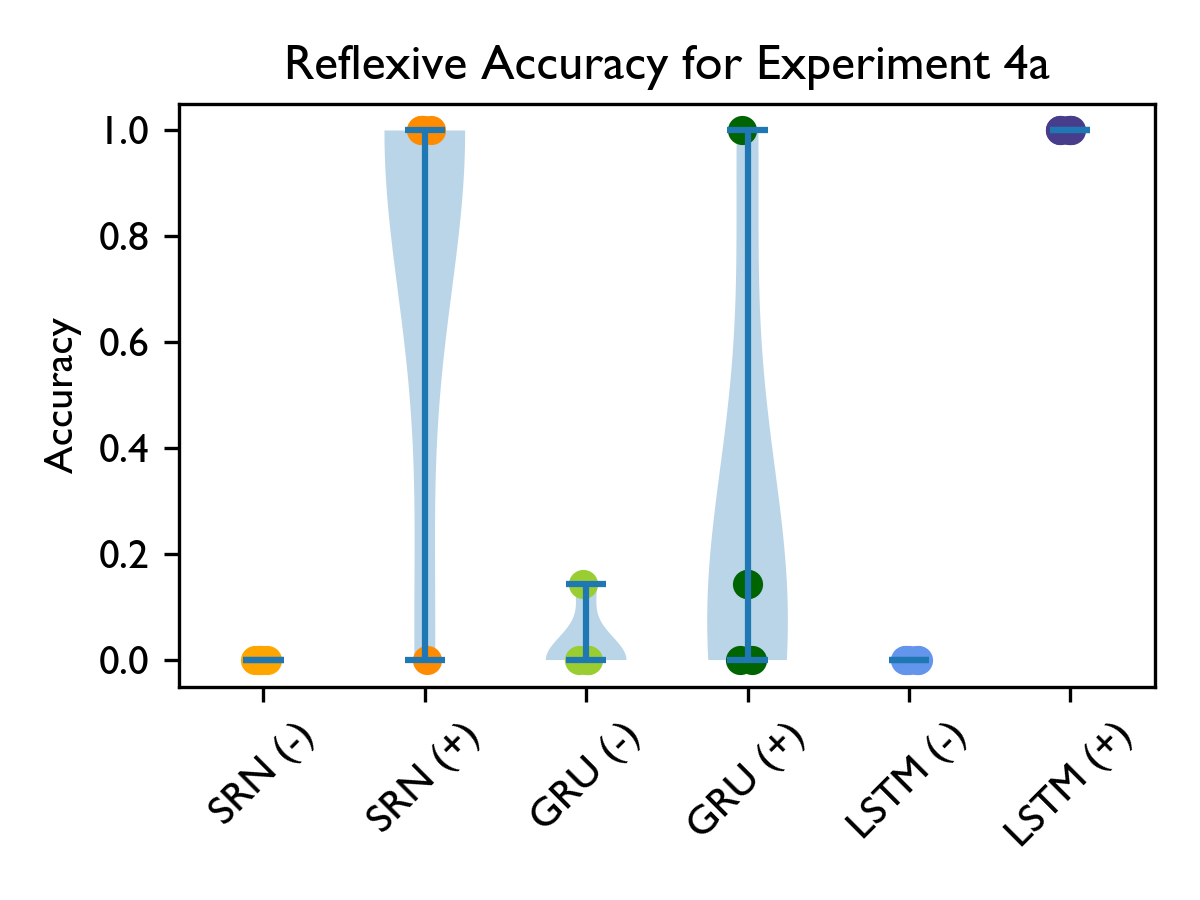}
    \hfill
    \includegraphics[width=0.45\columnwidth]{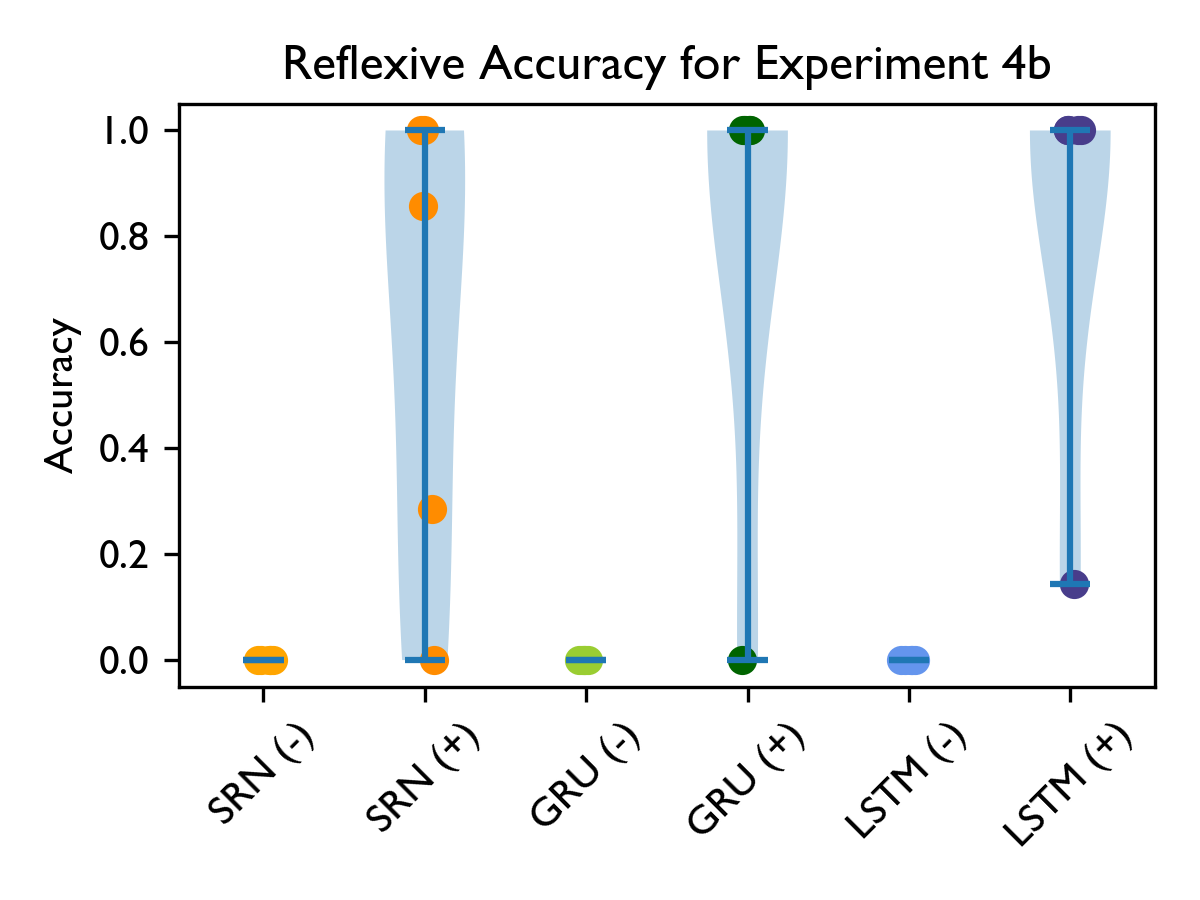}
    \vspace*{-1.5em}
    \caption{Mean accuracy on \lex{Alice}-reflexive sentences in Experiments~4a (left) and 4b (right).}
    \label{fig:four-violin}
\end{figure*}

The left plot in Figure~\ref{fig:four-violin} shows the reflexive generalization 
accuracy for the runs of the different architectures in the first variant of this 
experiment. Models without attention uniformly perform poorly across all recurrent 
unit types. With attention, performance is more variable: LSTMs perform at ceiling 
and SRNs do well for most random seeds, while GRUs perform poorly for most 
initializations with a single seed performing at ceiling. 
The top portion of Table~\ref{tab:four} contrasts the means of these results with the
generalization performance on transitives with \lex{Alice} subjects. 
Here again LSTMs without attention performed poorly while those with attention
did much worse on \lex{Alice}-transitives than on \lex{Alice}-reflexive 
sentences.
\begin{table*}
    \centering 
    \begin{tabularx}{\textwidth}{Xrrrrrr} \toprule
     \textit{Experiment 4a}   & SRN ($-$) & SRN ($+$) & GRU ($-$) & GRU ($+$) & LSTM ($-$) & LSTM ($+$) \\
        \cmidrule[0.5pt](lr{0.125em}){2-3}%
        \cmidrule[0.5pt](lr{0.125em}){4-5}%
        \cmidrule[0.5pt](lr{0.125em}){6-7}%
        \lex{Alice}-reflexive & 0.00 & 0.80 & 0.03 & 0.26 & 0.00 & \textbf{1.00} \\
        \lex{Alice}-subject (trans) & 0.02 & \textbf{0.83} & 0.04 & 0.29 & 0.03 & 0.28 \\
        \midrule
       \textit{Experiment 4b} & SRN ($-$) & SRN ($+$) & GRU ($-$) & GRU ($+$) & LSTM ($-$) & LSTM ($+$) \\ 
        \cmidrule[0.5pt](lr{0.125em}){2-3}%
        \cmidrule[0.5pt](lr{0.125em}){4-5}%
        \cmidrule[0.5pt](lr{0.125em}){6-7}%
        \lex{Alice}-reflexive & 0.00 & 0.63 & 0.00 & 0.80 & 0.00 & \textbf{0.83} \\
        \lex{Alice}-subject (trans) & 0.00 & 0.25 & 0.01 & \textbf{0.78} & 0.03 & 0.23 \\
        \lex{Alice}-subject (intrans) & 0.00 & 0.80 & 0.58 & 0.95 & 0.98 & \textbf{1.00} \\
        \bottomrule
    \end{tabularx}
    \caption{Mean accuracy on generalization sets for Experiments 4a and 4b.}
    \label{tab:four}
\end{table*}

This result at once highlights
the role that attention plays in learning this type of systematic generalization; attention appears to be necessary for recurrent architectures to generalize in this context.  The pattern of results also demonstrates a substantial effect of model architecture: 
attentive SRNs substantially outperform the more complex LSTM and
GRU architectures on generalization to \lex{Alice}-transitives, though this was not the case for reflexive sentences, where LSTMs showed a substantial advantage.

\paragraph{Experiment 4b}

The right plot in Figure~\ref{fig:four-violin} shows the impact of withholding  
\lex{Alice}-intransitive sentences from training. As before, models without
attention  fail on interpreting \lex{Alice}-reflexive
sentences. 
LSTMs and SRNs with attention perform nearly as well as in Experiment 4a, with some seeds performing at ceiling and a somewhat larger number than before failing to doing so. In contrast, the performance of attentive GRUs is improved in this context. The bottom of Table~\ref{tab:four} shows the mean generalization accuracy for transitive and intransitive sentences with \lex{Alice} subjects.  In some cases the transitive subject performance is as in Experiment 4a or worse, but in one case, namely attentive GRUs, it improves in this more difficult context, paralleling what we saw for reflexive generalization.

The reversal of GRU ($+$) and SRN ($+$) accuracies
better lines up with what we might expect given the complexity of the network 
architectures, with the more complex GRUs now outperforming the simpler
SRNs. These results also reinforce the connection observed in those from 
Experiment 4b on the effects of attention in generalization.





While withholding more information during training as we move from Experiment 4a to 4b might be 
expected to impair generalization for attentive GRUs, as it did for all other architectures, we in
fact see an increase in performance on \lex{Alice}-reflexive sentences. One possible explanation 
of this surprising result is that the attentive GRU networks in experiment 4a have learned from 
the training data a context-sensitive regularity concerning the distribution of the withheld name 
\lex{Alice}, namely that it  occurs only as the subject of intransitive verbs.  In Experiment 4b, 
however, the absence of evidence concerning the types of predicates with which \lex{Alice} may 
occur allows the network to fall back to a context-free generalization about \lex{Alice}, namely 
that it has the same distribution as the other names in the domain.  Note that this explanation is
possible only if the network treats intransitive and transitive subjects in a similar way.

\section{Experiment 5: What if nobody knows Alice?}

In the final experiment, we restrict the grammatical context in which \lex{Alice} appears by removing from the training data of Experiment 2 all instances of transitive sentences with \lex{Alice} in object position (but it is retained in subject position, apart from reflexive sentences).
In a second variant (Experiment 5b), we further restrict the training data to exclude all intransitive sentences with \lex{Alice} subjects. 
Although English, as a language with nominative-accusative alignment, treats
subjects of intransitives in a grammatically parallel fashion to subjects of
transitives, other languages (with ergative-absolutive alignment) treat intransitive subjects like transitive objects.  Though the word order of our synthetic language suggests nominative-accusative alignment, intransitive subjects have in common with transitive objects being the final argument in the logical form, which might lead to them being treated in similar fashion.

\subsection{Results}


\begin{figure*}[!t]
    \centering
    \includegraphics[width=0.45\columnwidth]{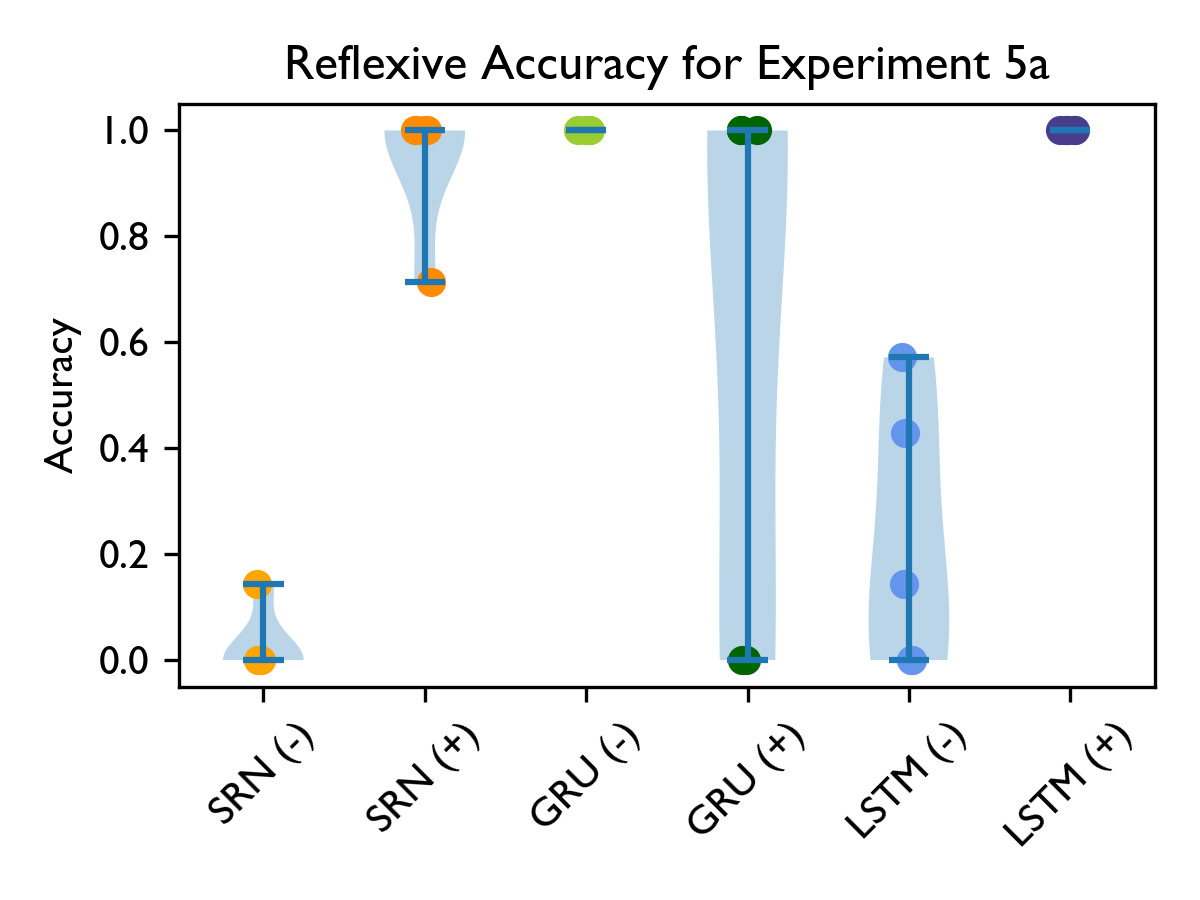}
    \hfill
    \includegraphics[width=0.45\columnwidth]{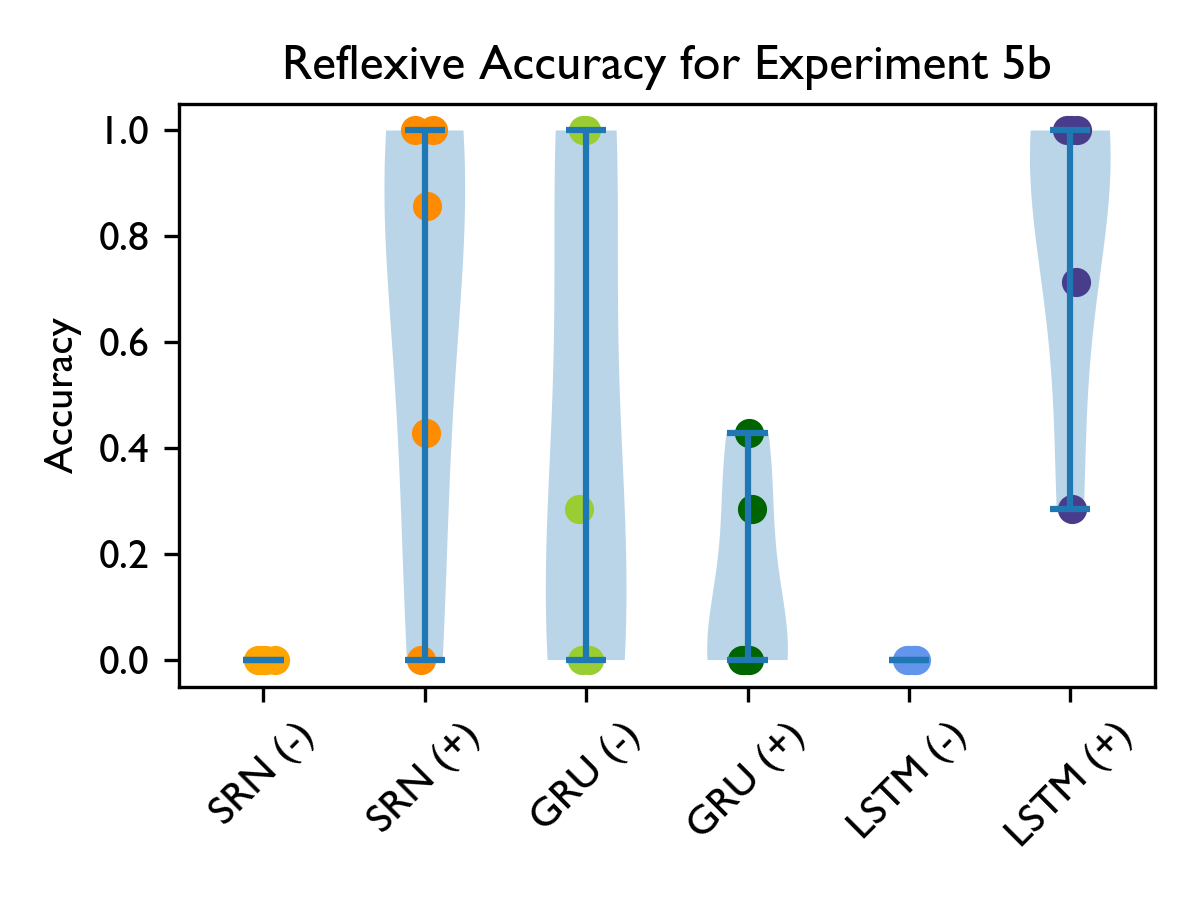}
       \vspace*{-1.5em}
    \caption{Mean accuracy on \lex{Alice}-reflexive sentences in Experiments~5a (left) and 5b (right).}
    \label{fig:five-violin}
\end{figure*}

\paragraph{Experiment 5a}
The left plot in Figure~\ref{fig:five-violin} shows  
reflexive generalization accuracy when the missing antecedent \lex{Alice} is withheld from transitive objects.
In contrast to the results in Experiment~4, the effect of attention is more varied here.
While SRNs and LSTMs without attention perform poorly, GRUs without attention perform well (for some seeds). As the top panel in Table~\ref{tab:five} shows, no models without attention performed well on 
sentences with \lex{Alice} in object position. For the models with attention, 
SRNs and LSTMs perforrmed uniformly well while the performance of GRUs was more mixed.  On \lex{Alice}-object sentences attentive SRNs again showed excellent performance, whereas the GRUs and LSTMs fared less well. At the same time, while GRUs with attention  outperformed GRUs without attention on \lex{Alice}-object sentences (25\% to 4\%), they
greatly underperformed them on the reflexive sentences (60\% to 98\%).


\begin{table*}[!bt]
    \centering
    \begin{tabularx}{\textwidth}{Xrrrrrr} \toprule
        \textit{Experiment 5a} & SRN ($-$) & SRN ($+$) & GRU ($-$) & GRU ($+$) & LSTM ($-$) & LSTM ($+$) \\ 
        \cmidrule[0.5pt](lr{0.125em}){2-3}%
        \cmidrule[0.5pt](lr{0.125em}){4-5}%
        \cmidrule[0.5pt](lr{0.125em}){6-7}%
        \lex{Alice}-reflexive & 0.03 & 0.94 & 0.98 & 0.60 & 0.23 & \textbf{1.00} \\
        \lex{Alice}-object & 0.00 & \textbf{0.97} & 0.04 & 0.25 & 0.04 & 0.37 \\ \midrule
         \textit{Experiment 5b} & SRN ($-$) & SRN ($+$) & GRU ($-$) & GRU ($+$) & LSTM ($-$) & LSTM ($+$) \\ 
        \cmidrule[0.5pt](lr{0.125em}){2-3}%
        \cmidrule[0.5pt](lr{0.125em}){4-5}%
        \cmidrule[0.5pt](lr{0.125em}){6-7}%
        \lex{Alice}-reflexive & 0.00 & 0.65 & 0.45 & 0.14 & 0.00 & \textbf{0.80} \\
        \lex{Alice}-object & 0.00 & \textbf{0.94} & 0.03 & 0.09 & 0.03 & 0.17 \\
        \lex{Alice}-subject (intrans) & 0.00 & 0.13 & 0.00 & 0.00 & 0.00 & \textbf{0.40} \\
        \bottomrule
    \end{tabularx}
    \caption{Mean accuracy on generalization sets for Experiments 5a and 5b.}
    \label{tab:five}
\end{table*}



\paragraph{Experiment 5b}

The right plots in Figure~\ref{fig:five-violin} shows the effects of further withholding
\lex{Alice}-intransitive sentences for \lex{Alice}-reflexive sentences. This manipulation has 
devastating effects on the performance of all models without attention. For models with 
attention, there is also a negative impact on reflexive generalization, but not as severe. As 
shown in the bottom portion of  Table~\ref{tab:five}, this manipulation has little impact on the 
network's performance on \lex{Alice}-object sentences, with SRNs with attention continuing to 
perform strongly and the other models performing less well.
GRUs continue to interact with attention in unusual ways. While they perform poorly on \lex{Alice}-object 
and \lex{Alice}-intransitive sentences with and without attention, inattentive GRUs continue to outperform attentive ones on reflexive sentences.

Overall, as in Experiment 4, LSTMs with attention show the highest accuracy
on the \lex{Alice}-reflexive sentences by a wide margin, while
SRNs with attention attain the best performance on 
\lex{Alice}-object sentences. Unlike in Experiment~4, withholding
the \lex{Alice}-intransitive sentences from training does not
yield any benefit for GRUs with attention in performance on the reflexive
set, in fact the opposite is true. This may be interpreted once again as evidence that GRUs are treating transitive and intransitive subjects as belonging to the same category. In Experiment 5a, \lex{Alice} occurs in both positions, leading the network to treat it as a subject like any other, and therefore potentially capable of serving as a subject of a reflexive. \lex{Alice}'s absence from object position does not impact the formation of this generalization. In Experiment 5b, on the other hand, where \lex{Alice} occurs only as a transitive subject, it leads the attentive GRU to treat it as name with a distinctive distribution, which impairs generalization to reflexive sentences.

\section{Conclusions}

Because of their abstract meaning, reflexive anaphora present a distinctive challenge for semantic parsing that had been thought to be beyond the capabilities of recurrent networks. The experiments described here demonstrate that this was incorrect. Sequence-to-sequence networks with a range of recurrent unit types are in fact capable of learning an interpretation of reflexive pronouns that generalizes to novel antecedents. Our results also show  that such generalization is nonetheless contingent on the appearance of the held-out  antecedent in a variety of syntactic positions as well as the diversity of antecedents providing support for the reflexive generalization. Additionally successful generalization depends on the  network architecture in ways that we do not fully understand. It is at present unknown whether the demands that any of these architecture impose on the learning environment for successful learning of reflexives are consistent with what children experience, but this could be explored with both corpus and experimental work.  Future work will also be necessary to elucidate the nature of the networks' representations of reflexive interpretation and to understand how they support lexical generalization (or not).  

The question we have explored here is related to, but distinct from, the issue of
systematicity \cite{FodorPylyshyn88,Hadley94}, according to which pieces of representations learned in distinct contexts can freely recombine.  This issue has been addressed  using sequence-to-sequence architectures  in recent work with the synthetic SCAN robot command interpretation dataset \citep{lake2018generalization} and on  language modeling \citep{kim-linzen-2020-COGS}, in both cases with limited success.  One aspect of the SCAN domain that is particularly relevant to reflexive interpretation is commands involving adverbial modifiers such as \lex{twice}.  Commands like \lex{jump twice} must be interpreted by duplicating the meaning of the verb, i.e., as \textsc{jump jump}, which is similar to what we require for the interpretation of the reflexive object, though in a way that does not require sensitivity to syntactic structure that we have not explored here.
Recently, \citet{lake2019compositional}, \citet{li-etal-2019-compositional} and \citet{Gordon2020Permutation} have proposed novel architectures that increase systematic behavior, and we look forward to exploring the degree to which these impact performance on reflexive interpretation. 

Our current work has focused 
exclusively on recurrent networks, ranging from  SRNs to GRUs and LSTMs. Recent work by \citet{transformer} shows that Transformer
networks  attain superior performance on a variety of sequence-to-sequence tasks while dispensing with recurrent units altogether.
Examining both the performance and training characteristics of Transformers will allow us
to compare the effects of attention and recurrence on the anaphora interpretation task. This is especially interesting given the impact that attention had
on performance in our experiments.

Finally, while our current experiments are revealing about the capacity of recurrent networks to learn generalizations about context-sensitive interpretation, there are nonetheless limited in a number of respects because of simplifications in the English fragment we use to create our synthetic data. Reflexives famously impose a structural requirement on their antecedents (c-command). In the following example, the reflexive's antecedent must be $\student$ and cannot be $\teacher$.
\ex
The student near the teacher sees herself  $\to$ $\see(\student, \student)$
\xe
We do not know whether the architectures that have succeed on our experiments would do similarly well if the relevant generalization required reference to (implicit) structure. Past work has explored the sensitivity of recurrent networks to hierarchical structure, with mixed results \cite{linzen-etal-2016,mccoy2020does}. In ongoing work, we are exploring this question by studying  more complex synthetic domains both with the kind of recurrent sequence-to-sequence network used here as well networks that explicitly encode or decode sentences in a hierarchical manner.   A second simplification concerns the distribution of reflexives themselves. English reflexives can appear in a broader range of syntactic environments apart from transitive objects \cite{storoshenko2008}. It would be of considerable interest to explore the reflexive interpretation in a naturalistic setting that incorporate this broader set of distributions.

 \section*{Acknowledgments}

For helpful comments and discussion of this work, we are grateful to Shayna Sragovicz, Noah Amsel,
Tal Linzen and the members of the Computational Linguistics at Yale (CLAY) and the JHU Computation and Psycholinguistics labs. This work has been supported in part by NSF grant BCS-1919321 and a Yale College Summer Experience Award.  Code for 
replicating these experiments can be found on the Computational Linguistics at the CLAY Lab GitHub \href{https://github.com/clay-lab/transductions}{\texttt{transductions}} and \href{https://github.com/clay-lab/logos}{\texttt{logos}} repositories.

\bibliography{anthology,acl2020}

\end{document}